\documentclass[conference]{IEEEtran}
\IEEEoverridecommandlockouts
\usepackage{cite}
\usepackage{amsmath,amssymb,amsfonts}
\usepackage{algorithmic}
\usepackage{graphicx}
\usepackage{textcomp}
\usepackage{xcolor}
\usepackage{bm}
\usepackage{caption}
\usepackage{subcaption}
\usepackage[hidelinks]{hyperref}
\usepackage[absolute]{textpos}
\def\BibTeX{{\rm B\kern-.05em{\sc i\kern-.025em b}\kern-.08em
    T\kern-.1667em\lower.7ex\hbox{E}\kern-.125emX}}
    
\usepackage[some]{background}
\SetBgScale{1}
\SetBgContents{\parbox{10cm}{%
  \Huge Draft:  \today\\[18cm]\rotatebox{180}{\Huge Draft:  \today}}}
\SetBgColor{red}
\SetBgAngle{270}
\SetBgOpacity{0.2}

\begin{document}

% To add the text above the title:
%\setlength{\TPHorizModule}{3.5cm}
%\setlength{\TPVertModule}{1cm}
%\begin{textblock}{5}(0.5,0.5)
%\noindent This work has been submitted to the IEEE for possible publication. Copyright may be transferred without notice, after which this version may no longer be accessible.
%\end{textblock}
%%%%%%%%%%%%%%%%%%%%

\title{Visual Material Characteristics Learning for Circular Healthcare\\
\thanks{This work has been conducted as part of the research project ‘Circular Economy for Small Medical Devices (ReMed)’, which is funded by the Engineering and Physical Sciences Research Council (EPSRC) of the UKRI (contract no: EP/W002566/1). The project funders were not directly involved in the writing of this article. For the purpose of open access, the author(s) has applied a Creative Commons Attribution (CC BY) license to any Accepted Manuscript version arising.}
\thanks{(\emph{Corresponding author: Federico Zocco})}
\thanks{The authors are with the Centre for Sustainable Manufacturing and Recycling Technologies (SMART), Wolfson School of Mechanical, Electrical and Manufacturing Engineering, Loughborough University, England, United Kingdom (e-mail: federico.zocco.fz@gmail.com, s.rahimifard@lboro.ac.uk)}
}

\author{\IEEEauthorblockN{Federico Zocco and Shahin Rahimifard}}
%\and

\maketitle

% Command to add the watermark in the title page specifying the version: 
%\BgThispage
%%%%%%

\begin{abstract}
The linear take-make-dispose paradigm at the foundations of our traditional economy is proving to be unsustainable due to waste pollution and material supply uncertainties. Hence, increasing the circularity of material flows is necessary. In this paper, we make a step towards circular healthcare by developing several vision systems targeting three main circular economy tasks: resources mapping and quantification, waste sorting, and disassembly. The performance of our systems demonstrates that representation-learning vision can improve the recovery chain, where autonomous systems are key enablers due to the contamination risks. We also published two fully-annotated datasets for image segmentation and for key-point tracking in disassembly operations of inhalers and glucose meters. The datasets and source code are publicly available\footnote{\url{https://github.com/fedezocco/MatVisionGluInh-PyTorch_TensorFlow}}.  
\end{abstract}

\section{Introduction} 
The traditional (i.e., linear) economy provides goods and services, but it relies on the extraction of non-renewable raw materials and it produces waste. Hence, the traditional supply chains alter the environment in a way that is unsustainable as demonstrated by the large accumulations of marine debris \cite{MarineDebris}, the large concentration of atmospheric carbon dioxide \cite{NASAcarbon}, and the uncertainties for the supplies of non-renewable materials in UK \cite{UKcritMat}, EU \cite{EUcritMat}, and US \cite{UScritMat}. Therefore, a transition from a linear to a circular economy (CE) is necessary. Note that, in this paper, the terms ``linear'' and ``circular'' refer to the material flows as outlined by Zocco et al. \cite{TMNpaper}.         

\begin{figure}
\includegraphics[width=0.48\textwidth]{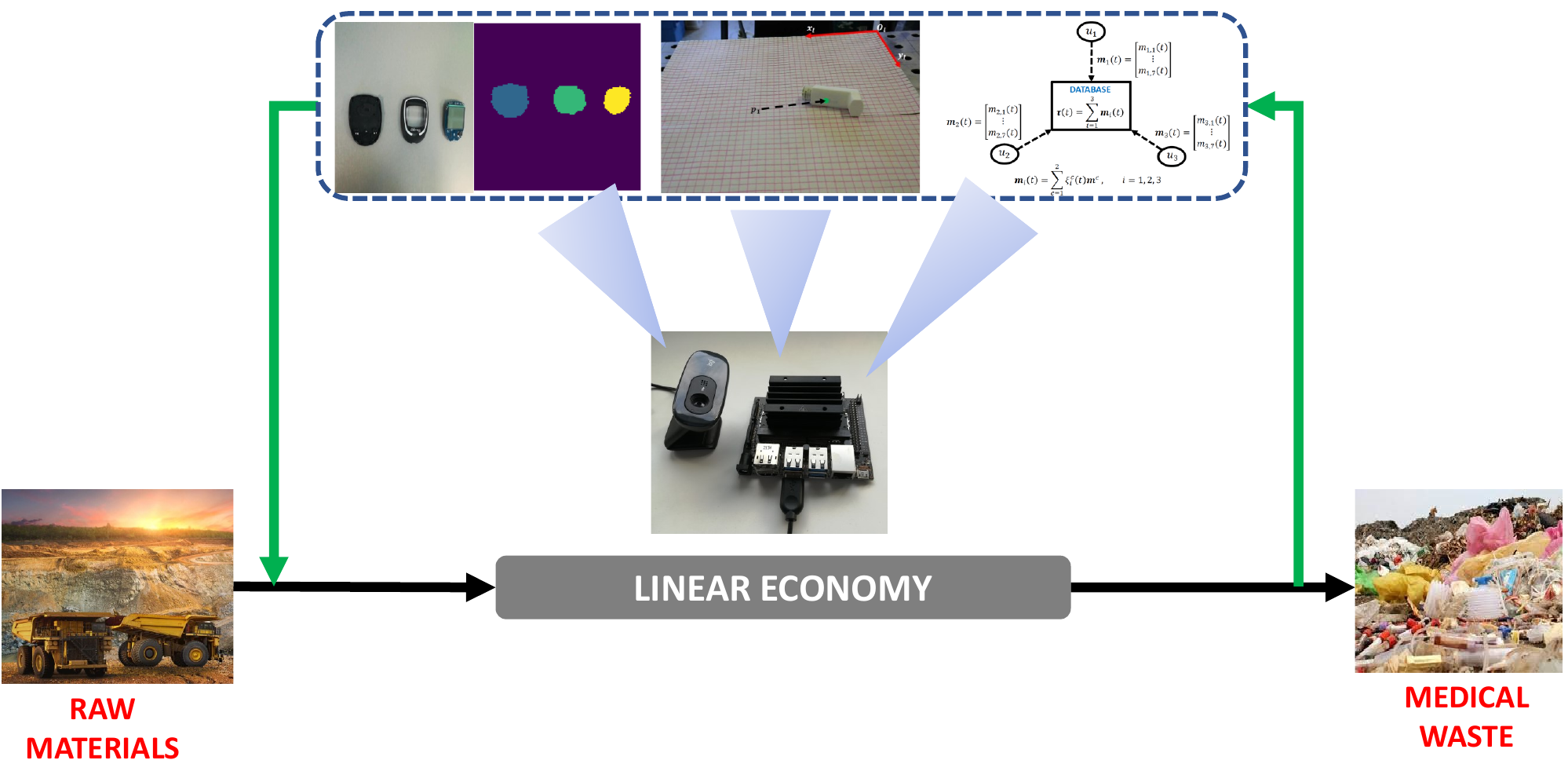}
\centering
\caption{High-level summary of the paper. The arrows indicate the material flows.}
\label{fig:forTitlePage}
\end{figure}

Healthcare is a large and essential sector of the economy of any country, hence its circularity is gaining importance \cite{EllenHealth,KPMGreport,transformHealth}. In UK, the NHS providers produce every year approximately 156,000 tonnes of clinical waste, which is equivalent to over 400 loaded jumbo jets \cite{NHSwaste}. The use of autonomous systems for handling end-of-first-use medical devices is promising for two main reasons: firstly, to reduce the risks of contamination due to the human contact with medical waste; and secondly, to reduce the costs of reprocessing large volumes of end-of-first-use devices. Hence, this paper focuses on computer vision (CV) for a circular healthcare as depicted in Fig. \ref{fig:forTitlePage}. While one type of glucose meter and inhaler have been considered as case-study products, it should be noted that the vision systems proposed here can be adapted to process other medical devices such as fluid bags, infusion pumps, syringes, laparoscopic scissors, along with their variants (e.g., to date, we are aware of at least 7 types of inhalers).   
% Figure to keep at the top right of the title page

\begin{figure*}
\centering
\includegraphics[width=0.90\textwidth]{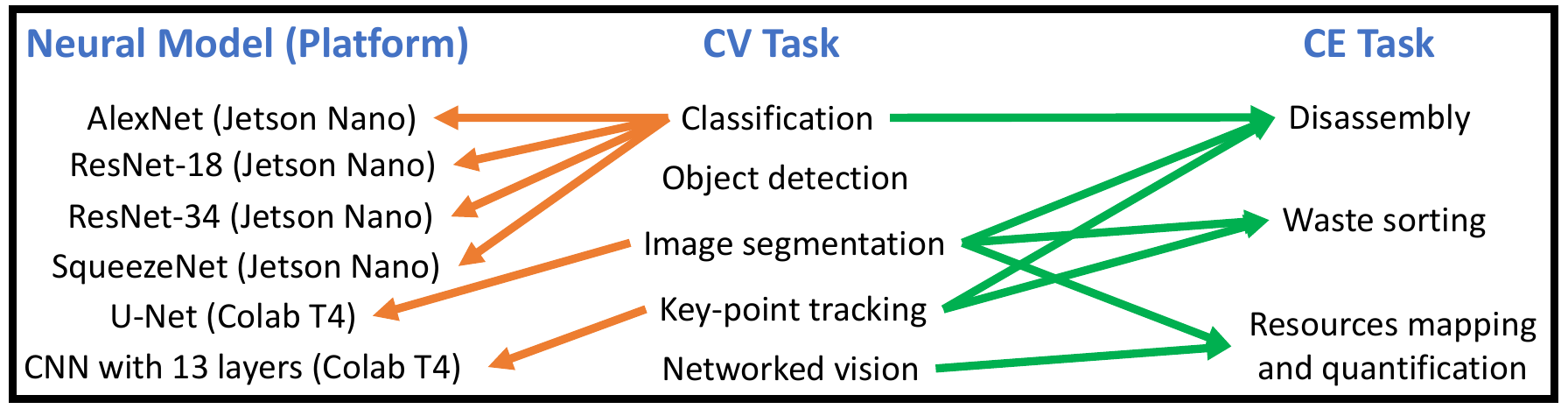}
\caption{Overview of tasks performed in this paper. Each computer vision (CV) task listed in the middle column is performed to address one or more circular economy (CE) tasks listed on the right using the neural model and GPU platforms listed on the left. We did not perform ``Object detection'', but it is mentioned for completeness.}
\label{fig:taskOverview}
\end{figure*}

\textbf{Our contributions:}
\begin{enumerate}
\item{We developed a set of vision systems for the automation of the following circular economy tasks: \emph{waste sorting}, \emph{disassembly}, and \emph{resources mapping and quantification} (our vision systems and circular economy tasks are summarized in Fig. \ref{fig:taskOverview}). By doing so, we also show that advanced computer vision has the potential to speed-up the transition from material flow linearity to circularity.}
\item{Our vision systems targeted two small medical devices whose recovery chain automation, to date, is minimal. Contamination risks and high costs make automation particularly suitable for the recovery chain of glucose meters and inhalers.}
\item{We annotated and published two datasets for circular economy applications, namely, GluMetF and InhPoints: GluMetF is for segmentation of glucose meter parts, while InhPoints is to estimate the coordinates of the pick points of an inhaler for robotic \emph{disassembly}.}
\end{enumerate}

A detailed overview of the tasks addressed in this paper is given in Fig. \ref{fig:taskOverview}. Our first CV task is ``Classification'', which is the simplest, yet most fundamental vision task; it could be used at the initial stage of robotic \emph{disassembly} to quickly select the sequence of actions optimized for a certain class (covered in Sections \ref{subsubsec:class} and \ref{subsec:class}). ``Object detection'' is included for completeness, but not performed in this paper. Hence, our second CV task is ``Image segmentation'', whose informative visual output makes it useful for the automation of any CE operations (covered in Sections \ref{subsubsec:segm} and \ref{subsec:segm}). Our third CV task is ``Key-point tracking'', which can be used in \emph{disassembly} and \emph{waste sorting} to provide to the robot the coordinates of specific points, e.g., the pick points (covered in Sections \ref{subsubsec:truck} and \ref{subsec:truck}). Our fourth and last CV task is ``Networked vision'', which is needed for autonomous \emph{resources mapping and quantification} (covered in Sections \ref{subsub:netVision} and \ref{sub:netVisionResults}).   

Note that the paper title mentions ``material characteristics learning'' because we focus on the natural resources to adopt a circular economy perspective. Specifically, we see an inhaler as the \emph{functionality} of its constituent materials; similarly, we see the \emph{position} of an inhaler as the position of its constituent materials; \emph{functionality} and \emph{position} are two examples of the characteristics of the materials learned by our vision systems.

The paper is organized as follows: Section \ref{sec:RelWork} covers the related work, Section \ref{sec:VisualMat} details the datasets, models, and training settings, then Section \ref{sec:Results} presents and analyses the results, and finally Section \ref{sec:Conclusions} concludes and discusses the future work.   
Throughout the paper, matrices and vectors are indicated with bold upper and lower case letters, respectively, while sets are indicated with calligraphic letters.

\section{Related Work}\label{sec:RelWork}
\subsection{Computer Vision for Disassembly}
In general, vision algorithms can be divided into two main categories: those based on hand-crafted features and those based on representation learning \cite{MMU}. The works \cite{forDisassembly3} and \cite{forDisassembly1} belong to the first category; the former focused on the removal of screws from the back of laptops, whereas the latter focused on the disassembling of electric vehicle motors. Our vision systems belong to the second category because of the higher accuracy it can reach \cite{SzeliskiBook}. This is in common with \cite{forDisassembly5,forDisassembly4,forDisassembly2}, which proposed neural object detection algorithms for detecting screws, and with \cite{forDisassembly6}, which performed both detection and segmentation of smart-phone components. Our algorithms perform classification, image segmentation, key-point tracking, and networked vision to further foster vision systems for tackling circular economy challenges (see Fig. \ref{fig:taskOverview}).

\subsection{Computer Vision for Waste Management} 
Vision systems along with public datasets have been proposed for litter recognition in water bodies to address the increasing waste accumulations \cite{CVwaste7,CVwaste3,CVwaste5}. Construction and demolition waste sorting was performed by Wang et al. \cite{CVwaste9}, while the system of Bai et al. \cite{CVwaste8} enabled their robot to detect garbage on grass before picking it. Recently, a large dataset for general waste segmentation has been published \cite{CVwaste6}, which highlights the complexity of the task from a vision perspective due to the extremely high shape (e.g., due to damages) and color (e.g., due to dirt) unpredictability that end-of-use items can have. Similarly, we provided a small dataset for segmentation as it is a particularly informative (but complex) vision task. In contrast, we focused on glucose meter parts and we also provided a second small dataset specifically for tracking pick-points, which could be used in both disassembling and waste sorting. Vision for robotic mixed waste sorting has been recently covered in \cite{CVwaste4}.

\subsection{Computer Vision for Resources Monitoring} 
One of the main methodologies to assess the circularity of materials is material flow analysis (MFA), which essentially consists of analyzing large amounts of historical data of material stocks and flows \cite{forMFA2,forMFA1,forMFA3}. Recently, Zocco et al. \cite{MMU} proposed (without implementation) the development of a network of vision systems for monitoring material stocks and flows to both improve the resources management and to provide input data to MFA studies. In this paper, we report a first numerical study of such a network (see Sections \ref{subsub:netVision} and \ref{sub:netVisionResults}). While both mass estimation \cite{massEstimation1,massEstimation2,massEstimation4,massEstimation5} and material recognition \cite{forMatRecognition1,forMatRecognition2,forMatRecognition3,forMatRecognition4} from images have been proposed to improve robotic manipulation, in this paper they are relevant for performing autonomous resources mapping and quantification (see Section 4 of \cite{MMU}): the former quantifies the amount of material, the latter specifies the type.

\section{Visual Material Characteristics Learning}\label{sec:VisualMat}
This section details our datasets, models, and training settings for visual characteristics learning of materials.

\subsection{Datasets}\label{sub:Datasets}
\subsubsection{GluMetF} It is a fully-annotated dataset for image segmentation of glucose meter parts. Specifically, three main parts are segmented: the back case, the front case, and the printed circuit board (PCB). While the first two parts are plastic-based, the latter is made of tiny quantities of different metals, some of them with long-term supply uncertainties \cite{MMU}. The ``F'' in the dataset name indicates that we labeled considering the \emph{functionality} of the materials: their functionality is either the back case, the front case, or the PCB. The dataset has 280 samples for training, 80 samples for validation, and 40 samples for testing. The images were extracted from videos recorded with a HUAWEI P8 lite 2017 and they were annotated using Computer Vision Annotation Tool by Intel \cite{CVAT}. GluMetF has the format of the Cityscapes dataset \cite{Cityscapes}.

\subsubsection{InhPoints}
\begin{figure}
\begin{subfigure}{0.15\textwidth}
  \centering
  \includegraphics[width=1\textwidth]{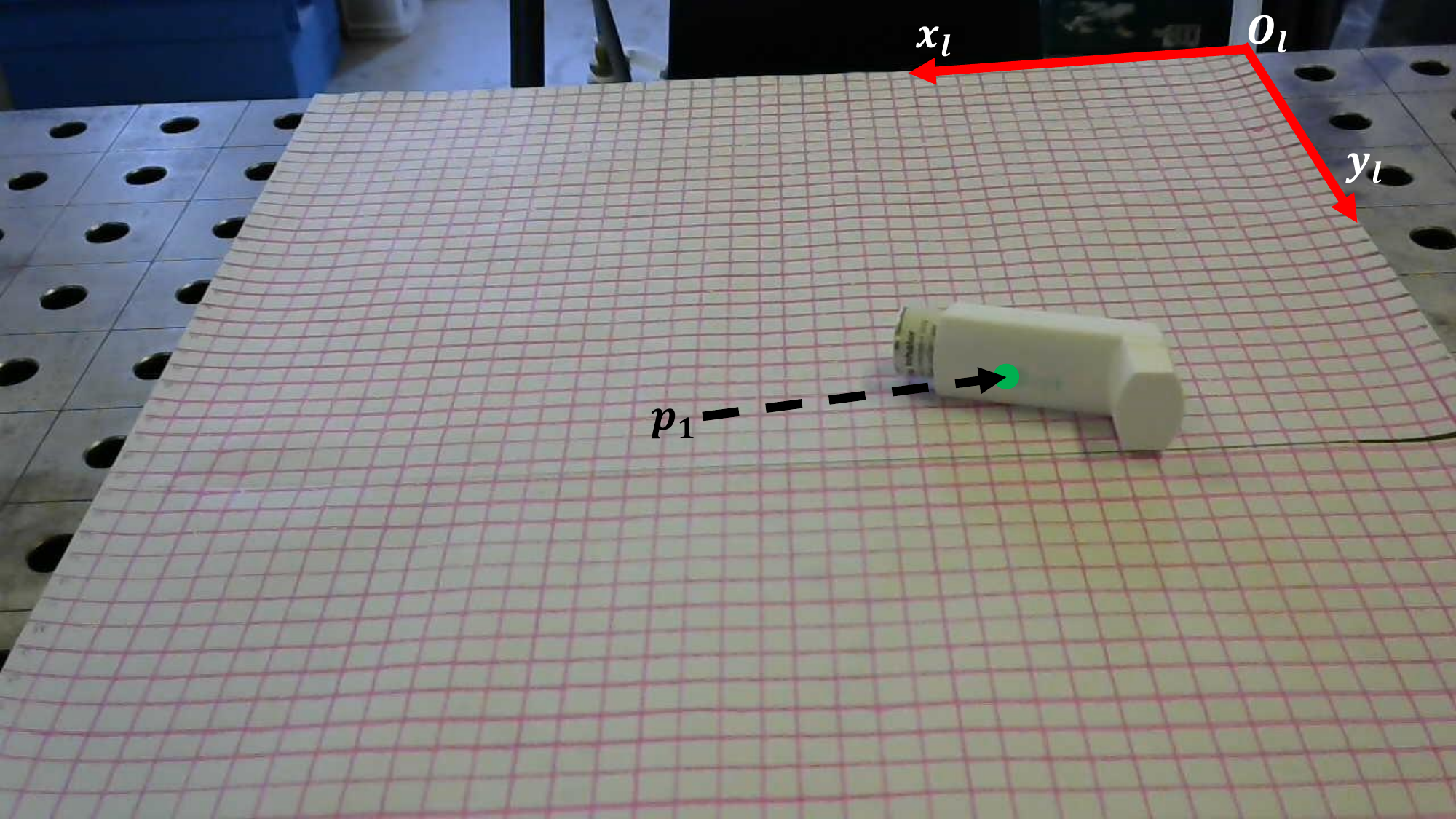}
  \caption{Local frame and pick point $p_1$}
  \label{fig:inhTracking-p1}
\end{subfigure}
\begin{subfigure}{0.15\textwidth}
  \centering
  \includegraphics[width=1\textwidth]{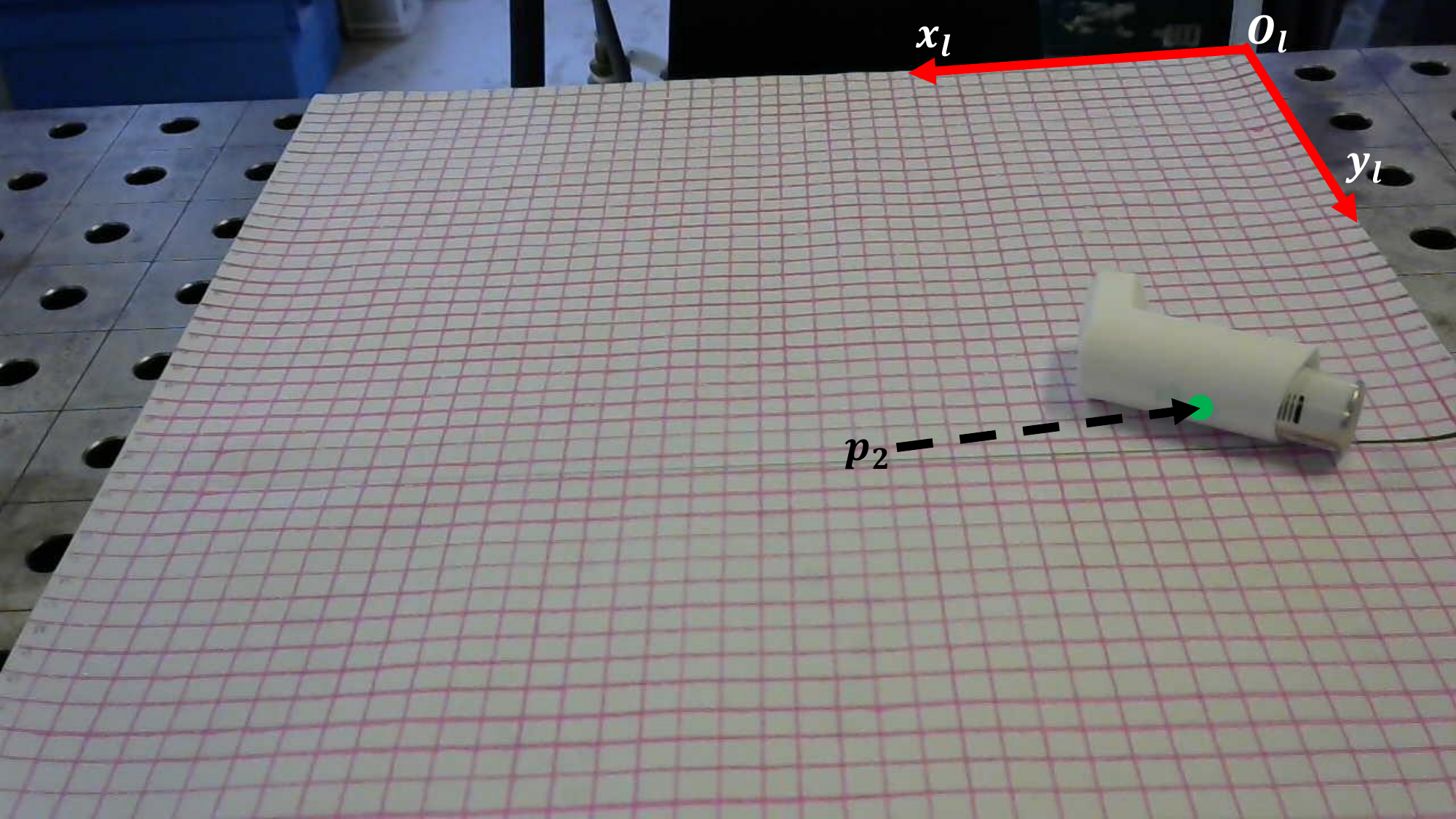}
  \caption{Local frame and pick point $p_2$}
  \label{fig:inhTracking-p2}
  %\vspace{5mm}
\end{subfigure} 
\begin{subfigure}{0.15\textwidth}
  \centering
  \includegraphics[width=1.1\textwidth]{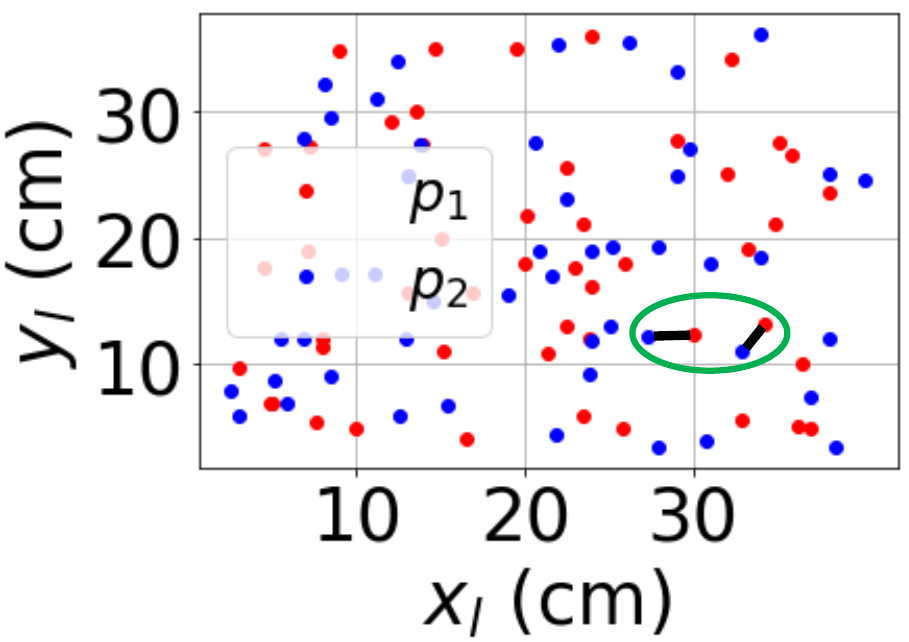}
  \caption{Targets}
  \label{fig:inhTracking-targets}
  %\vspace{5mm}
\end{subfigure}
\caption{Setup for the tracking of the inhaler pick points to begin the disassembly process.}
\label{fig:inhTracking}
\end{figure}

Biological systems use vision to locate the objects. If no visual information is used, a robotic arm programmed to pick an object expects the object to be located in a known position and with a known orientation. The InhPoints dataset eliminates the requirement of this a-priori knowledge by using the visual tracking of image key-points. The data collection of InhPoints was performed considering the initial stage of the robotic disassembly of an inhaler, in which the arm needs to know the coordinates of the pick points $p_1$ and $p_2$ as shown in Fig. \ref{fig:inhTracking-p1} and Fig. \ref{fig:inhTracking-p2} with respect to a local frame with origin and axes $O_\text{l}$, $x_\text{l}$, and $y_\text{l}$, respectively. This can be stated as a regression problem solved through supervised learning, in which the images captured by an RGB camera oriented towards a robotic bench are processed by a neural network that is trained to predict the targets in Fig. \ref{fig:inhTracking-targets}. The targets are coordinate pairs ($x_{\text{l},i}; y_{\text{l},i}$) of the pick point $p_i$, $i \in \{1, 2\}$. Note that each sample pair ($x_{\text{l},1}; y_{\text{l},1}$) corresponds to a sample pair ($x_{\text{l},2}; y_{\text{l},2}$) as they are two points of the same rigid body, i.e., the inhaler. To clarify, two examples are shown in Fig. \ref{fig:inhTracking-targets} (inside the green circle), where the black link between points indicates the inhaler body. 
Note also that we placed a grid between the inhaler and the bench for two reasons (see Fig. \ref{fig:inhTracking-p1}): firstly, it allowed the manual annotation of the local coordinates $(x_{\text{l},i}; y_{\text{l},i})$ of $p_i$ during the data collection; and secondly, it should help the vision algorithm to detect variations of position and orientation of the inhaler in the same way it helps the detection by a human observer. 

Thus, InhPoints consists of 50 images related to a target vector $\bm{t}_{{\text{l}}_\text{l}} = [x_{\text{l},1}, y_{\text{l},1}, x_{\text{l},2}, y_{\text{l},2}]^\top$, where $\bm{t}_{{\text{l}}_\text{l}} \in \mathbb{R}^4$ indicates the target vector with tail in $O_\text{l}$ and written with respect to the local frame. In general, a robot has its own frame with origin and axes $O_\text{r}$, $x_\text{r}$, and $y_\text{r}$, respectively. If $\bm{t}_{{\text{l}}_\text{l}} \in \mathbb{R}^4$ (given by the vision system) is split into $\bm{t}^1_{{\text{l}}_\text{l}} = [x_{\text{l},1}, y_{\text{l},1}, h]^\top$ and $\bm{t}^2_{{\text{l}}_\text{l}} = [x_{\text{l},2}, y_{\text{l},2}, h]^\top$, then, $\bm{t}^i_{{\text{l}}_\text{l}} \in \mathbb{R}^3$ can be written with respect to the robot frame as     
\begin{equation}\label{eq:changeOfCoordinates}
\bm{t}^i_{{\text{r}}_\text{r}} = \bm{d}_{{\text{r},\text{l}}_\text{r}} + \bm{R}_{\text{r},\text{l}} \bm{t}^i_{{\text{l}}_\text{l}}, \quad i \in \{1, 2\},
\end{equation}
where $\bm{R}_{\text{r},\text{l}} \in \mathbb{R}^{3 \times 3}$ is the rotation matrix of the local frame with respect to the robot frame, $\bm{d}_{{\text{r},\text{l}}_\text{r}} \in \mathbb{R}^{3}$ is the translation vector having the tail in $O_\text{r}$ and the head in $O_\text{l}$, and $h$ is a constant value that indicates the coordinate $z_\text{l}$ of the local frame normal to the bench. The quantities $\bm{R}_{\text{r},\text{l}}$, $\bm{d}_{{\text{r},\text{l}}_\text{r}}$ are known since they are defined by the position of the local frame, that is, the grid on the bench, while $h$ is known from the geometry of the inhaler. Hence, (\ref{eq:changeOfCoordinates}) provides to the robot the position of the pick points $p_1$ and $p_2$ in the robot frame, i.e., $\bm{t}^1_{{\text{r}}_\text{r}}$ and $\bm{t}^2_{{\text{r}}_\text{r}}$, respectively. 

Note that, even if a small dataset, InhPoints introduces to the complexity of visual key-point tracking in disassembly operations: a complete vision system should work with different types of inhalers, which may also have different end-of-first-use conditions that are very difficult to predict, e.g., a missing part or a fracture.

\subsection{Models and Trainings}\label{sub:ModAndTrain}
\subsubsection{Classification}\label{subsubsec:class}
We performed the simplest vision tasks, that is, classification, on an NVIDIA Jetson Nano with 128-core Maxwell GPU, quad-core ARM A57 CPU with clock speed of 1.43 GHz, and 4 GB 64-bit LPDDR4 memory with speed of 25.6 GB/s. Both training and inference were executed on it. 

We designed our classifiers following an NVIDIA Deep Learning Institute tutorial \cite{DLItutorial}, which is based on PyTorch; 100 images of glucose meters and 100 images of inhalers were collected with the camera pointing at a robotic bench for a binary classification. All images were down-sized to $224 \times 224$ and normalized. Then, we fine-tuned a pre-trained AlexNet \cite{AlexNet}, a pre-trained ResNet-18 \cite{ResNets}, a pre-trained ResNet-34 \cite{ResNets}, and a pre-trained SqueezeNet \cite{SqueezeNet} for 5 epochs with batch size of 8, Adam optimizer, and the cross-entropy loss. We compared the trained models in terms of accuracy and floating-point operations (FLOPs).

\subsubsection{Image segmentation}\label{subsubsec:segm}
The standard classification considered above is image-wise, hence it does not provide information about the shapes, number and positions of the target objects. This information is necessary for an effective robotic disassembly so that the robot can approach the object with the appropriate speed and stiffness. Such an information can be provided by image segmentation, which performs a pixel-wise classification.

We designed the image segmenter leveraging the neural architecture provided in the TensorFlow tutorial \cite{TFsegmentation}. The whole network is a U-Net \cite{Unet}, whose encoder is a pre-trained MobileNetV2 \cite{Mobilenetv2} and whose decoder is a series of 4 upsample blocks whose implementation is based on pix2pix \cite{pix2pix}. Each image was down-sized to $128 \times 128$ before the neural processing. The training was performed on Google Colaboratory (aka Colab) with an NVIDIA T4 GPU with 12.7 GB of RAM; only the decoder was trained, while the encoder was frozen. We trained the model for 50 epochs with a batch size of 4, Adam as the optimizer with a learning rate of 0.0002, and with the cross-entropy loss. We trained the image segmenter on GluMetF dataset for the recognition of the main glucose meter parts, that is, the front case, the back case, and the PCB.

\subsubsection{Key-Point Tracking}\label{subsubsec:truck}
The task of key-point tracking was performed by training a CNN on the InhPoints dataset (see Section \ref{sub:Datasets}). Hence, in this case, the key-points were the inhaler pick points $p_1$ and $p_2$ as shown in Fig. \ref{fig:inhTracking}. Specifically, we designed a CNN with 13 layers to estimate $\bm{t}_{{\text{l}}_\text{l}}$ from RGB images. The training was performed on Colab with an NVIDIA T4 GPU with 12.7 GB of RAM. The whole network was trained from scratch. The model was trained for 200 epochs with a batch size of 2, with Adam as the optimizer, and with the mean-squared-error loss.

\subsubsection{Networked Vision}\label{subsub:netVision}
Let us consider a vision system which is part of a network of multiple vision systems. We will refer to it as a ``unit'' of the networked vision. This unit detects a product whose material composition is of interest and, therefore, it is stored in the unit memory. Then, the unit detecting the target product communicates the material information to a central database. If the location of the unit is known (e.g., via the GPS), the location of the materials is known as well. If a network of multiple units is deployed to cover a target urban area, material mapping and quantification could be performed to improve the management of resources \cite{MMU}.

\begin{figure}
\centering
\includegraphics[width=0.48\textwidth]{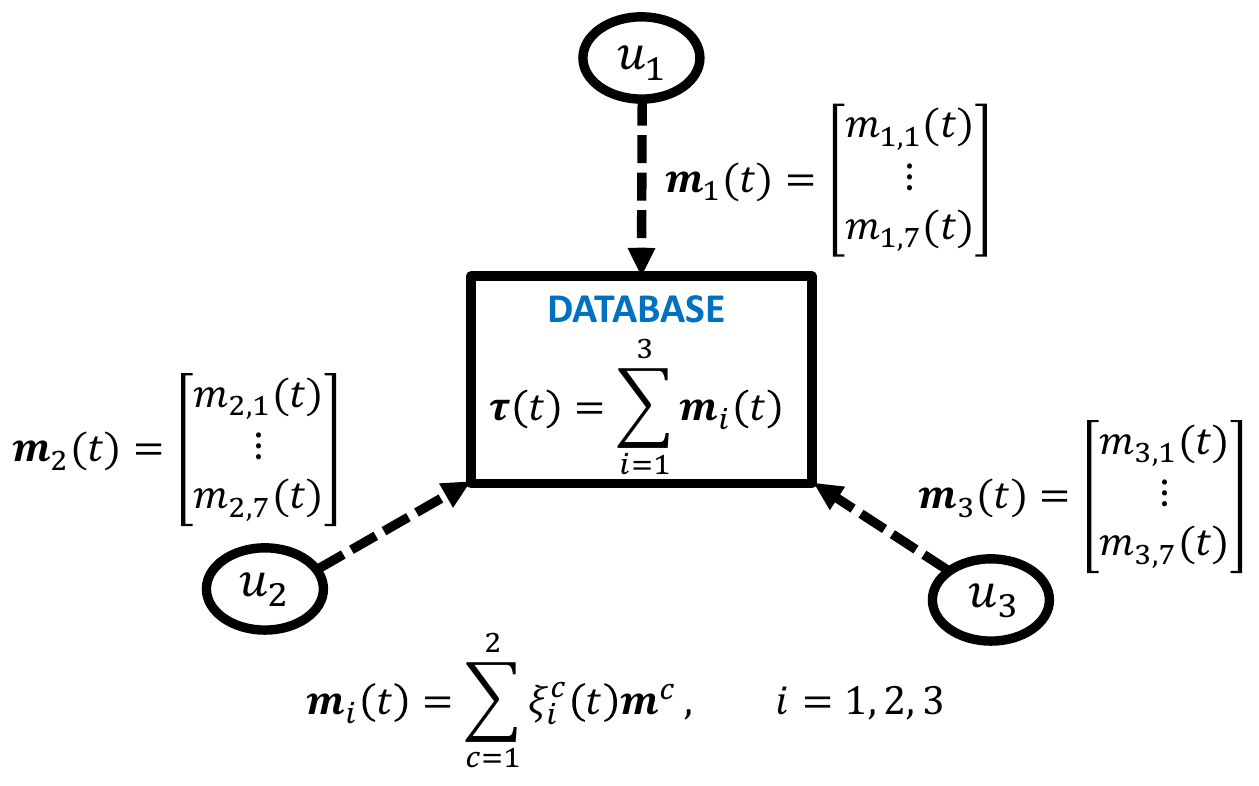}
\caption{Vision system for autonomous material mapping and quantification in the case of $s = 3$, $\psi = 7$, and $q = 2$.}
\label{fig:networkedVision}
\end{figure}

Now, let us assume the ideal case in which each unit performs a given recognition task with no errors, e.g., the binary classification covered in Section \ref{subsubsec:class} in which glucose meters and inhalers are detected. Be $\mathcal{U} = \{u_1, \dots, u_i, \dots, u_s\}$ a set of networked vision units, $C = \{1, 2, \dots, c, \dots, q\}$ the set of object-classes detected by the $i$-th unit $u_i$, $\mathcal{M}_{\text{mon}} = \{1, 2, \dots, j, \dots, \psi\}$ the set of materials being monitored by the vision network $\mathcal{U}$ with $|\mathcal{M}_{\text{mon}}| = \psi$, and $\mathcal{M}_{\text{all}}$ is the set of all the constituent materials of the detected object-classes $c \in C$ (e.g., the constituent materials of a glucose meter if ``glucose meter'' is a class $c \in C$). Hence, $\mathcal{M}_{\text{mon}} \subseteq \mathcal{M}_{\text{all}}$. Be also $\bm{m}^c = [m^c_1, \dots, m^c_j, \dots, m^c_\psi]^\top \in \mathbb{R}^\psi$ the masses of the materials detected for the object-class $c$ and be $\bm{m}_i(t) = [m_{i,1}(t), \dots, m_{i,j}(t), \dots, m_{i,\psi}(t)]^\top \in \mathbb{R}^\psi$ the masses of materials detected by the $i$-th unit. The latter is computed as
\begin{equation}
\bm{m}_i(t) = \sum_{c = 1}^{q} \xi_i^c(t) \bm{m}^c,
\end{equation} 
where 
\begin{equation}\label{eq:rectFunction}
\xi_i^c(t) = \text{rect}\left(\frac{t - T^c_{\text{p},i}}{T^c_{\text{d},i}}\right),
\end{equation}
\begin{equation}
\text{rect}\left(\frac{t}{T}\right) = 
	\begin{cases}
      1, & |t/T| < 1/2 \\
      1/2, & |t/T| = 1/2 \\
      0, & \text{otherwise}
	\end{cases}  
\end{equation}
is the rectangular function centered in $t = 0$ \cite{Vitetta-TdS}, $T^c_{\text{p},i}$ is the center of the shifted rectangular impulse, and $T^c_{\text{d},i}$ is the duration of the rectangular impulse. The rectangular function in (\ref{eq:rectFunction}) models the detection of an object-class $c \in \mathcal{C}$ performed by the unit $u_i$, which starts in $t = T^c_{\text{p},i} - \frac{T^c_{\text{d},i}}{2}$ and has a duration of $T^c_{\text{d},i}$. The material information from all the units is collected in the central database as
\begin{equation}
\bm{\tau}(t) = [\tau_1(t), \dots, \tau_j(t), \dots, \tau_\psi(t)]^\top = \sum^s_{i = 1} \bm{m}_i(t).
\end{equation}
Figure \ref{fig:networkedVision} provides a graphical representation of the whole system in the case of $s = 3$, $\psi = 7$, and $q = 2$. 

Note that, for simplicity of exposition, we assume that $\text{length}(\bm{m}^c) = \text{length}(\bm{m}_i) = \psi$, where $\text{length}(v)$ takes the number of elements of $v$. In the most general case, $\text{length}(\bm{m}^c) \leq \text{length}(\bm{m}_i)$, i.e., the number of materials detected for the object-class $c$ are less than or equal to the number of materials detected by $u_i$.

\section{Results and Analysis}\label{sec:Results}
This section reports and discusses the results achieved by performing the four vision tasks covered previously in Section \ref{sub:ModAndTrain} for material circularity. 
\subsection{Classification}\label{subsec:class}
\begin{figure}
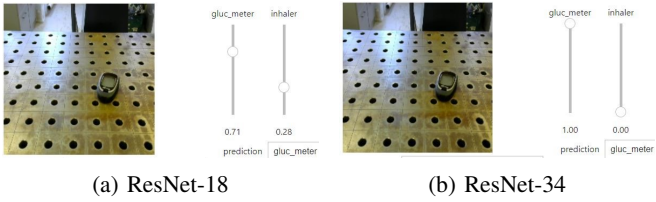

\begin{subfigure}{0.24\textwidth}
  \centering
  \includegraphics[trim={3cm 0cm 3cm 0}, width=0.5\textwidth]{Figures/resnet18\_qualitative}
  \caption{ResNet-18}
  \label{fig:class-ResNet-18}
  %\vspace{5mm}
\end{subfigure} 
\begin{subfigure}{0.24\textwidth}
  \centering
  \includegraphics[trim={3cm 0cm 3cm 0}, width=0.5\textwidth]{Figures/resnet34\_qualitative}
  \caption{ResNet-34}
  \label{fig:class-ResNet-34}
  %\vspace{5mm}
\end{subfigure}
\caption{Qualitative performance of the best classifiers running on an NVIDIA Jetson Nano processing real-time frames acquired on a robot bench.}
\label{fig:class-qualitative}
\end{figure}

A quantitative analysis of the trained classifiers is reported in Table \ref{tab:class-quantitative} in terms of accuracy and computational complexity. The most accurate is ResNet-34 (score: 1.00) followed by its smaller variant ResNet-18 (score: 0.95), then AlexNet and SqueezeNet are the least accurate with a score of 0.50. Note that a score of 0.50 means that the model has a probability of 0.5 to make the right prediction. Since the classification is binary, the trained AlexNet and SqueezeNet perform as good as a classifier that makes a random choice. The ranking of the four classifiers in terms of FLOPs is the same as for the accuracy: ResNet-34 is the most demanding (3.7B), while AlexNet and SqueezeNet are the least demanding (0.7B and 0.8B, respectively). This indicates that ResNet-34 has the largest inference time whereas AlexNet is the fastest. 
\begin{table}
\caption{Accuracy and FLOPs of the classifiers trained and tested on an NVIDIA Jetson Nano. Corresponding qualitative analysis is shown in Fig. \ref{fig:class-qualitative}.}
\centering
\begin{tabular}{c c c c c} 
 \hline
& AlexNet & ResNet-18 & ResNet-34 & SqueezeNet \\  
 \hline
Accuracy & 0.50 & 0.95 & 1.00 & 0.50\\  
FLOPs & 0.7B & 1.8B & 3.7B & 0.8B\\
 \hline
\end{tabular}
\label{tab:class-quantitative}
\end{table}
A qualitative analysis of the performance of the best classifiers is shown in Fig. \ref{fig:class-qualitative}: an unseen frame is given as input to each model and the probability of each class (i.e., either glucose meter or inhaler) is indicated by two vertical cursors. In agreement with Table \ref{tab:class-quantitative}, ResNet-34 is the most accurate as it predicts the correct class with probability of 1.0, followed by ResNet-18 with probability of 0.71, and finally AlexNet and SqueezeNet with probability of 0.5.     
Considering these results, the preferred models are ResNet-18 and ResNet-34: the former has a smaller latency gained at the cost of some accuracy, while the latter gives the best accuracy at the cost of 1.9B extra FLOPs compared to the former.

\subsection{Image Segmentation}\label{subsec:segm}
\begin{figure*}[t!]
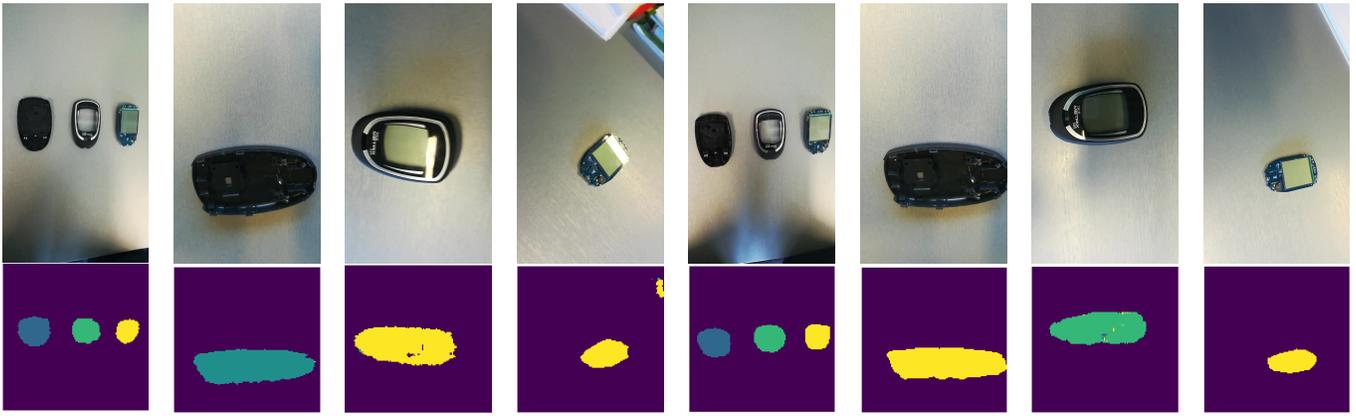

\begin{subfigure}{0.119\textwidth}
  \centering
  \includegraphics[width=0.9\textwidth]{Figures/all\_input\_1}
  \label{fig:segm-all-in-1}
\end{subfigure}
\begin{subfigure}{0.119\textwidth}
  \centering
  \includegraphics[width=0.9\textwidth]{Figures/backcase\_input\_1}
  \label{fig:segm-bck-in-1}
  %\vspace{5mm}
\end{subfigure} 
\begin{subfigure}{0.119\textwidth}
  \centering
  \includegraphics[width=0.9\textwidth]{Figures/frontcase\_input\_1}
  \label{fig:segm-frt-in-1}
  %\vspace{5mm}
\end{subfigure}
\begin{subfigure}{0.119\textwidth}
  \centering
  \includegraphics[width=0.9\textwidth]{Figures/pcb\_input\_1}
  \label{fig:segm-pcb-in-1}
  %\vspace{5mm}
  \end{subfigure}
  \begin{subfigure}{0.119\textwidth}
  \centering
  \includegraphics[width=0.9\textwidth]{Figures/all\_input\_2}
  \label{fig:segm-all-in-2}
\end{subfigure}
\begin{subfigure}{0.119\textwidth}
  \centering
  \includegraphics[width=0.9\textwidth]{Figures/backcase\_input\_2}
  \label{fig:segm-bck-in-2}
  %\vspace{5mm}
\end{subfigure} 
\begin{subfigure}{0.119\textwidth}
  \centering
  \includegraphics[width=0.9\textwidth]{Figures/frontcase\_input\_2}
  \label{fig:segm-frt-in-2}
  %\vspace{5mm}
\end{subfigure}
\begin{subfigure}{0.119\textwidth}
  \centering
  \includegraphics[width=0.9\textwidth]{Figures/pcb\_input\_2}
  \label{fig:segm-pcb-in-2}
  %\vspace{5mm}
\end{subfigure} \\
\begin{subfigure}{0.119\textwidth}
  \centering
  \includegraphics[width=0.9\textwidth]{Figures/all\_output\_1}
  \label{fig:segm-all-out-1}
\end{subfigure}
\begin{subfigure}{0.119\textwidth}
  \centering
  \includegraphics[width=0.9\textwidth]{Figures/backcase\_output\_1}
  \label{fig:segm-bck-out-1}
  %\vspace{5mm}
\end{subfigure} 
\begin{subfigure}{0.119\textwidth}
  \centering
  \includegraphics[width=0.9\textwidth]{Figures/frontcase\_output\_1}
  \label{fig:segm-frt-out-1}
  %\vspace{5mm}
\end{subfigure}
\begin{subfigure}{0.119\textwidth}
  \centering
  \includegraphics[width=0.9\textwidth]{Figures/pcb\_output\_1}
  \label{fig:segm-pcb-out-1}
  %\vspace{5mm}
\end{subfigure}
\begin{subfigure}{0.119\textwidth}
  \centering
  \includegraphics[width=0.9\textwidth]{Figures/all\_output\_2}
  \label{fig:segm-all-out-2}
\end{subfigure}
\begin{subfigure}{0.119\textwidth}
  \centering
  \includegraphics[width=0.9\textwidth]{Figures/backcase\_output\_2}
  \label{fig:segm-bck-out-2}
  %\vspace{5mm}
\end{subfigure} 
\begin{subfigure}{0.119\textwidth}
  \centering
  \includegraphics[width=0.9\textwidth]{Figures/frontcase\_output\_2}
  \label{fig:segm-frt-out-2}
  %\vspace{5mm}
\end{subfigure}
\begin{subfigure}{0.119\textwidth}
  \centering
  \includegraphics[width=0.9\textwidth]{Figures/pcb\_output\_2}
  \label{fig:segm-pcb-out-2}
  %\vspace{5mm}
\end{subfigure}
\caption{Qualitative performance analysis of image segmentation performed on test samples of the GluMetF dataset for the recognition of glucose meter parts. Top row: model input; bottom row: model output. The corresponding quantitative results are: accuracy of 0.997 and validation accuracy of 0.988.}
\label{fig:segm-qualitative}
\end{figure*}

Qualitative segmentation results on GluMetF dataset are shown in Fig. \ref{fig:segm-qualitative}, while the corresponding quantitative results are: accuracy of 0.997 and validation accuracy of 0.988.      
The top row of Fig. \ref{fig:segm-qualitative} shows the frame depicted by an RGB camera, which is the input to the segmentation model; the bottom row is the corresponding output produced by the model. The output has a smaller size than the input because the input has been down-sized before the neural processing in order to fit into the available RAM (12.7 GB). Overall, the glucose meter parts are segmented from the background in all the images and the shapes are predicted correctly. This shows clearly the advantage of image segmentation over standard classification: the segmented image can be used by a robotic arm to get information about the position and class of the object, which is analogous to the role of vision in biological systems. A more careful analysis of Fig. \ref{fig:segm-qualitative} shows that the segmentation is not perfect since the predicted classes, i.e., the colors, are not always consistent. For example, the first segmented image on the left-side shows three different colors, which is correct because there are three different objects in the input image. In particular, the blue is used for the back case. However, the second segmented image does not contain the blue, but it contains a color in between the blue of the back case and the green of the front case. Also, compare the third segmented image with the seventh one: the former misclassifies the front case as a PCB (in yellow), whereas the latter identifies the correct class (in green). The PCB is classified correctly as visible in the fourth and the eighth segmented images. 

Given that the validation accuracy after training is very high (i.e., 0.988), but such an high accuracy is not reached with the unseen samples of Fig. \ref{fig:segm-qualitative}, it seems that the model is partially overfitting the data. The model generalization could be improved by adding more annotated images to GluMetF and, then, repeating the training.

\subsection{Key-Point Tracking}\label{subsec:truck}
After the training, the mean and standard deviation of the error in predicting the targets in Fig. \ref{fig:inhTracking-targets} are 0.09 cm and 0.06 cm, respectively. These small values of the error should be read by considering two factors that may lead to higher errors when deploying the trained model: firstly, these error values are with respect to the training set, whereas unseen samples such as those processed in real-time during disassembly will be predicted with an accuracy that depends on the model generalization; and secondly, the target-coordinates of $p_1$ and $p_2$ were annotated by a human that was reading the projections of each point onto the grid we placed on the bench indicating $z_\text{l} = 0$, whereas $p_1$ and $p_2$ have $z_\text{l} = c \neq 0$ (Fig. \ref{fig:inhTracking-p1}); these annotation method has certainly resulted in some inaccuracies, which subsequently were introduced into the model weights during the training process.

\subsection{Networked Vision}\label{sub:netVisionResults}
This section shows a numerical study for the case $s = 2$, $\psi = 3$, and $q = 2$. To clarify the application of this material mapping and quantification system, assume that the first material stock being monitored by the network is plastic, i.e., $\tau_1(t)$, the second material stock is glass, i.e., $\tau_2(t)$, and the third material stock is gold, i.e., $\tau_3(t)$ (this network monitors $\psi = 3$ materials). Further values used for this study are: $T^1_{\text{p},1} = 60$ s, $T^2_{\text{p},1} = 70$ s, $T^1_{\text{p},2} = 80$ s, $T^2_{\text{p},2} = 90$ s, $T^1_{\text{d},1} = T^2_{\text{d},1} = T^1_{\text{d},2} = T^2_{\text{d},2} = 80$ s, $m^1_1 = m^2_1 = 1$ kg, $m^1_2 = m^2_2 = 2$ kg, and $m^1_3 = m^2_3 = 3$ kg. 

\begin{figure}
\begin{subfigure}{0.15\textwidth}
  \centering
  \includegraphics[width=1\textwidth]{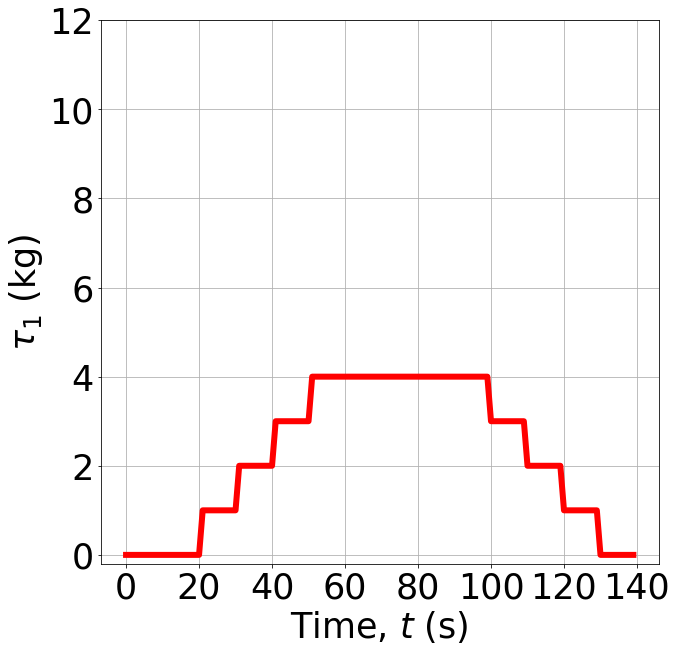}
  \caption{$\tau_1(t)$}
  \label{fig:monitoredMat1}
\end{subfigure}
\begin{subfigure}{0.15\textwidth}
  \centering
  \includegraphics[width=1\textwidth]{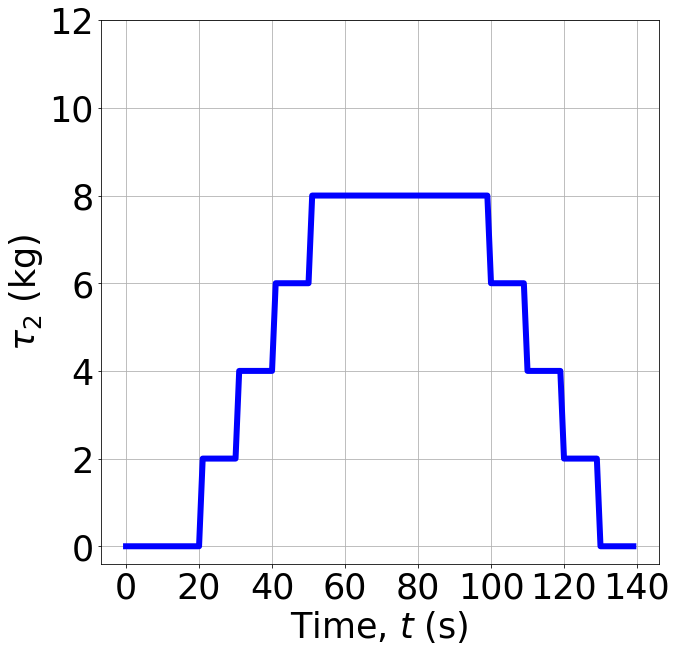}
  \caption{$\tau_2(t)$}
  \label{fig:monitoredMat2}
  %\vspace{5mm}
\end{subfigure} 
\begin{subfigure}{0.15\textwidth}
  \centering
  \includegraphics[width=1\textwidth]{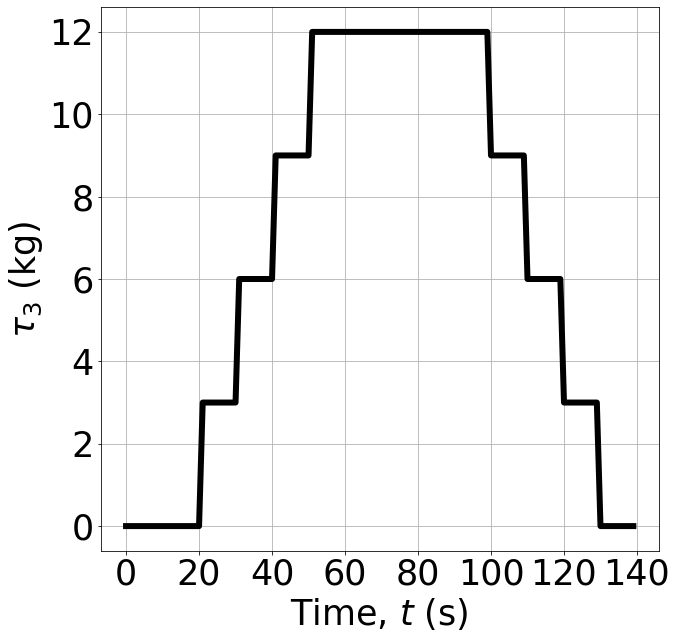}
  \caption{$\tau_3(t)$}
  \label{fig:monitoredMat3}
  %\vspace{5mm}
\end{subfigure}
\caption{The three material stocks monitored in the numerical study of material mapping and quantification.}
\label{fig:monitoredMat}
\end{figure}

The dynamics of the material stocks are shown in Fig. \ref{fig:monitoredMat}. 
All the three stocks $\tau_{j}(t)|_{j = 1, 2, 3}$ show variations in the same times because $T^c_{\text{p},i}$ and $T^c_{\text{d},i}$ are independent of the material type $j$; specifically, the stock variations are in $t = 20, 30, 40, 50, 100, 110, 120, 130$ s. The difference between the three stocks is in their magnitude: since $m^c_1 = 1$ kg $\forall c \in \mathcal{C}$, a new detection adds 1 kg to the first stock $\tau_1(t)$ (see $t = 20, 30, 40, 50$ s in Fig. \ref{fig:monitoredMat1}), while the end of a detection removes 1 kg from $\tau_1(t)$ (see $t = 100, 110, 120, 130$ s in Fig. \ref{fig:monitoredMat1}). In contrast, any new (or lost) detection adds (or removes) 2 kg to the second stock $\tau_2(t)$ because $m^c_2 = 2$ kg $\forall c \in \mathcal{C}$ (see Fig. \ref{fig:monitoredMat2}). Similarly, any new (or lost) detection adds (or removes) 3 kg to the third stock $\tau_3(t)$ because $m^c_3 = 3$ kg $\forall c \in \mathcal{C}$ (see Fig. \ref{fig:monitoredMat3}). Note also that the detection of class $c = 1$ by the unit $u_1$ starts in $t = 20$ s because $T^1_{\text{p},1} - T^1_{\text{d},1}/2 = 20$ s and it ends in $T^1_{\text{p},1} + T^1_{\text{d},1}/2 = 100$ s. Similarly, the detection of class $c = 2$ by the unit $u_1$ starts in $T^2_{\text{p},1} - T^2_{\text{d},1}/2 = 30$ s and ends in $t = 110$ s, the detection of class $c = 1$ by the unit $u_2$ starts in $t = 40$ s and ends in $t = 120$ s, and finally, the detection of class $c = 2$ by the unit $u_2$ starts in $t = 50$ s and ends in $t = 130$ s.        

The complexity of the whole system is regulated by three main parameters: the number of units $s$, the number of classes $q$, and the number of monitored materials $\psi$. To measure the material distribution inside a wide urban area, it would be required a network with values of $s$, $q$, and $\psi$ higher than the ones used in this study. In addition, further complexity will follow from removing the assumption that each unit classifies the objects with no errors. Since any misclassification propagates into the measure of stock $\bm{\tau}(t) \in \mathbb{R}^\psi$, the accuracy of the whole system is heavily affected by the accuracy of each unit.

\section{Conclusions}\label{sec:Conclusions}
%Summary of results and future work
In this paper, we developed several vision systems for enhancing the circularity of healthcare considering glucose meters and inhalers as case studies. Specifically, we considered three CE tasks: \emph{waste sorting}, \emph{disassembly} for repairing and recycling, and \emph{resources mapping and quantification}.

% CV task 1:
Among the four considered classifiers for \emph{disassembly}, the preferred models are ResNet-18 and ResNet-34. 
% CV task 2:
The proposed image segmenter for part recognition of glucose meters performs very well on training and validation sets, whereas it shows some misclassifications on unseen samples. An improvement of the accuracy would very likely be achieved by extending the size of the training set (currently with 280 samples) and, then, repeating the training for an higher number of epochs (50 were used here). This will be future work along with the implementation of the trained segmenter on the Jetson Nano to analyze the system latency during real-time operations.     

% CV task 3:
The vision-based regression model we developed for tracking the pick points of an inhaler showed very small errors: mean and standard deviation of 0.09 cm and 0.06 cm in predicting the training samples, respectively. A future work in this direction will be to combine this tracking system with the image segmenter mentioned above: the less-computationally-demanding tracking model runs when the gripper is far from the object, whereas the more-computationally-demanding segmentation model runs in the proximity of the object to use the shape information for optimizing the grasping. 

% CV task 4:  
Our numerical study of vision-enabled \emph{material mapping and quantification} illustrated the system functional principles. In particular, the complexity is determined by $\psi$, $s$, and $q$, which in this paper were set as 3, 2, and 2, respectively. Future work in this direction will be looking at increasing the values of these three parameters to analyze more realistic scenarios, removing the assumption of no misclassifications, and designing a first prototype of the whole system.  

Another future work will be to adapt these vision systems to process other types of glucose meters and inhalers, and also other medical devices such as infusion pumps, fluid bags, and syringes.

\section*{Acknowledgment}
The authors gratefully thank the ReMed Team (\url{https://www.remed.uk/team}) for the valuable discussions and Berk Calli at Worcester Polytechnic Institute for the useful comments. 
%The preferred spelling of the word ``acknowledgment'' in America is without 
%an ``e'' after the ``g''. Avoid the stilted expression ``one of us (R. B. 
%G.) thanks $\ldots$''. Instead, try ``R. B. G. thanks$\ldots$''. Put sponsor 
%acknowledgments in the unnumbered footnote on the first page.

%%%%% CLEAR DOUBLE PAGE!
\newpage{\pagestyle{empty}\cleardoublepage}

\end{document}